\documentclass{bmvc2k}
\usepackage{graphicx}
\usepackage{amsmath,amssymb} 
\usepackage{color}

\usepackage{times}
\usepackage{epsfig}
\usepackage{booktabs}
\usepackage{array}
\usepackage{rotating}

\title{In Defense of Single-column Networks for
Crowd Counting}
\addauthor{Ze Wang$^{*1,}$}{ze.w@buaa.edu.cn}{2}
\addauthor{Zehao Xiao$^{*1,}$}{zhxiao@buaa.edu.cn}{2}
\addauthor{Kai Xie}{kxie_INEE@163.com}{4}
\addauthor{Qiang Qiu}{qiang.qiu@duke.edu}{3}
\addauthor{Xiantong Zhen$^{1,}$}{zhenxt@buaa.edu.cn}{2}
\addauthor{Xianbin Cao$^{1,}$}{xbcao@buaa.edu.cn}{2}

\addinstitution{
 School of Electronic and Information Engineering,\\
 Beihang University,\\
 Beijing, China
}
\addinstitution{
 The Key Laboratory of Advanced Technologies for Near Space Information Systems,\\
 Ministry of Industry and Information Technology of China,\\
 Beijing, China
}
\addinstitution{
 Duke University,\\
 Durham, NC, USA
}
\addinstitution{
 Institute of North Electronic Equipment,\\
 Beijing, China
}

\runninghead{WANG \MakeLowercase{et al.}}{In Defense of Single-column Networks for
Crowd Counting}

\begin{document}

\maketitle
\begin{abstract}
Crowd counting usually addressed by density estimation becomes an increasingly important topic in computer vision due to its widespread applications in video surveillance, urban planning, and intelligence gathering. However, it is essentially a challenging task because of the greatly varied sizes of objects, coupled with severe occlusions and vague appearance of extremely small individuals. Existing methods heavily rely on multi-column learning architectures to extract multi-scale features, which however suffer from heavy computational cost, especially undesired for crowd counting. In this paper, we propose the single-column counting network (SCNet) for efficient crowd counting without relying on multi-column networks. SCNet consists of residual fusion modules (RFMs) for multi-scale feature extraction, a pyramid pooling module (PPM) for information fusion, and a sub-pixel convolutional module (SPCM) followed by a bilinear upsampling layer for resolution recovery. Those proposed modules enable our SCNet to fully capture multi-scale features in a compact single-column architecture and estimate high-resolution density map in an efficient way. In addition, we provide a principled paradigm for density map generation and data augmentation for training, which shows further improved performance. Extensive experiments on three benchmark datasets show that our SCNet delivers new state-of-the-art performance and surpasses previous methods by large margins, which demonstrates the great effectiveness of SCNet as a single-column network for crowd counting.
\end{abstract}

\section{Introduction}
\label{sec:intro}

Counting the number of people by estimating their density distribution from crowd images has attracted increasing attention because of its wide range of applications, such as safety monitoring, disaster management, public spaces design, and intelligence gathering \cite{sindagi2017survey}, especially in the congested scenes like arenas, shopping malls, and airports. However, it is not a trivial task due to great challenges caused by occlusion, clutter scene, irregular object distribution, non-uniform object scale, and inconstant perspective and background.

Recently, in an attempt to deal with those challenges, CNN based methods have been developed for crowd counting and density estimation \cite{walach2016learning,zhang2016single,marsden2016fully,sam2017switching,sindagi2017generating}, among which CNNs with a multi-column architecture, referred as multi-column CNN (MCNN) \cite{zhang2016single} were extensively studied. MCNN typically employs multi-branch sub-networks with filters of different sizes to extract multi-scale features for addressing the issue of various individual sizes. Existing methods mainly follow the multi-column architecture that shares the similar topology of multi-branch sub-networks. However, networks with the multi-column architecture usually introduce heavy computational overhead, being nontrivial to optimize \cite{sindagi2017survey}, especially when the network goes deeper, which therefore makes them unfavorable for the task of crowd counting.

In this paper, we propose the single-column counting network (SCNet) for crowd counting and density estimation, which establishes a simple but effective network without relying on the multi-scale architecture.
Our SCNet consists of several conjunctive modules which are designed for efficient crowd counting. Specifically, residual fusion modules (RFMs), composed of several nested dilated layers and short-cut connections, are stacked for multi-scale features extraction; a pyramid pooling module (PPM) is deployed to fuse hierarchically contextual information and forms the entire feature encoder for the generation of the semantic feature map in conjunction with the RFMs; a sub-pixel convolutional module (SPCM) with a bilinear interpolation operation is used to decode the semantic feature map for the high-resolution density estimation, and provides a parameter-free way for resolution recovery without compromising accuracy.

Moreover, we provide a principled paradigm for density map generation and data augmentation, based on which we introduce online sampling and multi-scale training to further enhance the overall performance.

In general, we can summarize our major contributions in the following three aspects:
\begin{itemize}

	

	\item We propose single-column counting network (SCNet) for crowd counting and density estimation with a simple, easy-to-implement architecture which achieves competitive and even better performance compared to the multi-column counterparts.

	\item We design the residual fusion modules and the pyramid pooling module for capturing multi-scale features to handle the great variation of object sizes; and we adopt the sub-pixel convolutional module for feature resolution recovery and density map generation in an efficient, nonparametric way.

	\item We conduct a throughout study on the experimental settings and data preparations in crowd counting and provide a principled paradigm to fully utilizing the limited training data.
	
\end{itemize}

Extensive experiments on three public benchmark datasets show that our SCNet can achieve state-of-the-art performance, surpassing most previous methods, which demonstrates its great effectiveness as a single-column network for crowd counting.

\begin{figure}[t]
	\begin{center}
		\includegraphics[width=1.0\linewidth]{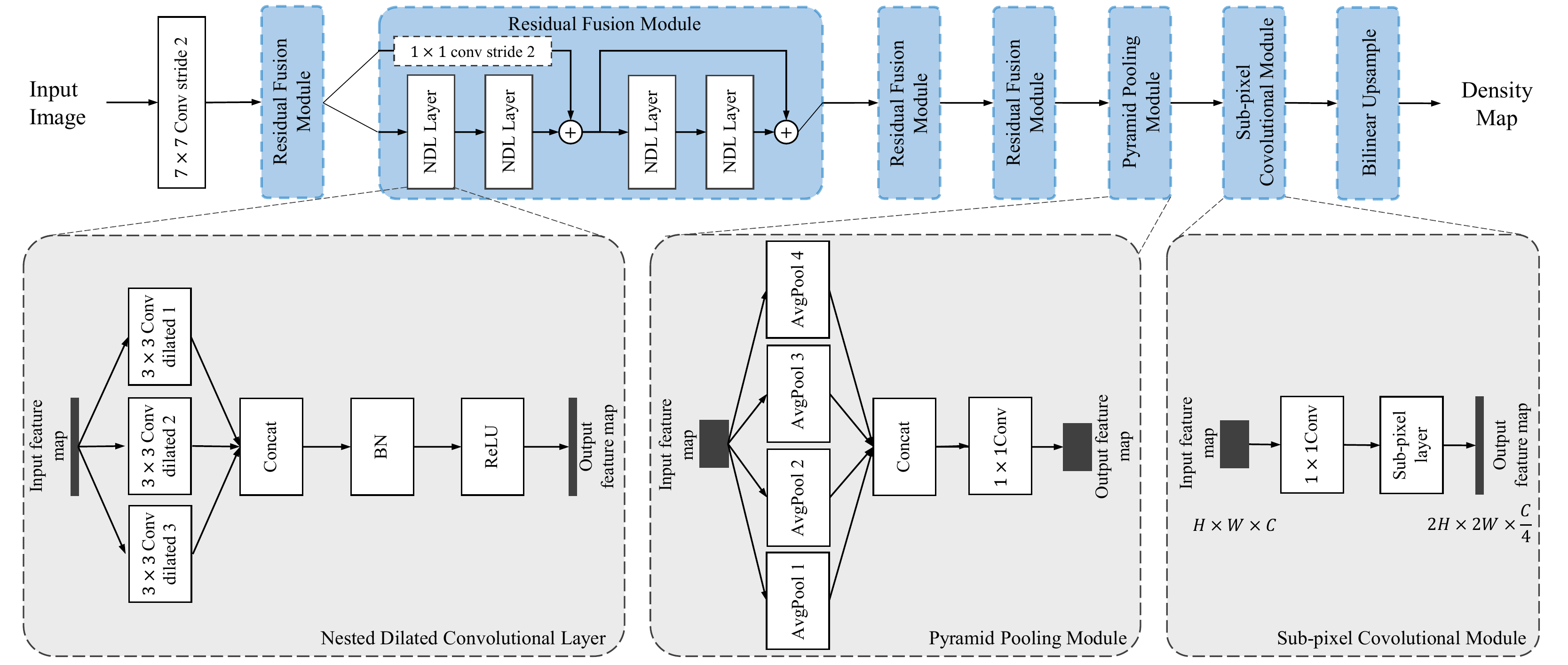}
	\end{center}
	\caption{The architecture of our Single-column Counting Net (SCNet). 
	}
	\label{fig:arc}
\end{figure}

\section{Related Work}
We describe the related work of crowd counting in two parts as in \cite{sindagi2017survey}: traditional approaches and CNN-based approaches.

Various methods have been proposed to address the problem of crowd counting. According to \cite{loy2013crowd}, the traditional approaches can be roughly divided into three categories: detection-based approaches, regression-based approaches, and density estimation-based approaches.

Most of the early work on crowd counting used sliding window detectors to detect people and count the number of them \cite{dalal2005histograms,leibe2005pedestrian,enzweiler2009monocular}. These methods extracted features of the whole pedestrian to train their classifiers and achieved successful results in low-density crowd scenes \cite{viola2004robust,dalal2005histograms}.
Going a step further, for better performance in high-density scenes, researchers adopted part-based detection methods that detect particular body parts rather than the whole body to estimate the people count \cite{felzenszwalb2010object,wu2007detection}.

Although the part-based detection methods lightened the problem of occlusion, they behaved poor performance in extremely dense crowd scenes and high background clutter scenes. Researchers then tried to use regression-based approaches to overcome these challenges \cite{chan2008privacy,chan2009bayesian,chen2013cumulative,chen2012feature,idrees2013multi}.
The regression-based approaches learn mappings between features extracted from images and the number of people in these images \cite{chan2008privacy,paragios2001mrf,chen2012feature}.

While the regression-based methods addressed the problems of occlusion and background clutter well, most of them ignored the important spatial information \cite{sindagi2017survey}. Lempitsky et al. \cite{lempitsky2010learning} proposed to learn a linear mapping between local patch features and corresponding object density maps to make full use of the spatial information. Differently, Pham et al. \cite{pham2015count} proposed to learn a non-linear one due to the difficulty of learning a linear mapping.

Inspired by the success of CNN in a large number of computer vision tasks \cite{zhen2018multi,zhen2018multitarget,miao2018direct}, researchers began to use CNN-based approaches to generate density maps and count the number of interest-objects \cite{walach2016learning,zhang2016single,marsden2016fully,sam2017switching,sindagi2017generating}.

Walach et al. \cite{walach2016learning} proposed a CNN-based method with layered boosting and selective sampling. In contrast to this patch-based training method, Shang et al. \cite{shang2016end} used an end-to-end CNN method that took the entire image as input and outputted the total crowd count. Boominathan et al. \cite{boominathan2016crowdnet} presented a fully convolutional network that combined a deep network and a shallow network to predict the density map while addressing scale variations across images.

Motivated by the multi-column networks for image classification \cite{ciregan2012multi}, the MCNN proposed by Zhang et al. \cite{zhang2016single} generated the density map by merging multi-scale features extracted by networks with different receptive fields. Similarly, Onoro et al. \cite{onoro2016towards} developed a scale aware counting model named Hydra CNN to estimate object densities. However, Marsden et al. \cite{marsden2016fully} proposed a single column fully convolutional network after observing the optimization difficulties and complicated calculations of earlier scale aware methods.

More recently, Sam et al. \cite{sam2017switching} proposed a switching CNN architecture that smartly selected the most suitable regressor for the particular input patch. Sindagi et al. \cite{sindagi2017cnn} developed a cascaded CNN network that applied the high-level prior to promote the prediction performance. The Contextual Pyramid CNN \cite{sindagi2017generating} extracted global and local contextual information by CNN networks and utilized the contextual information to achieve lower count error and improve the quality of density maps.
These CNN-based methods achieved the state-of-the-art performance in crowd counting.

\section{Single-column Counting Network}

Our SCNet consists of four residual fusion modules (RFM), a pyramid pooling module (PPM), and a sub-pixel convolutional module (SPCM), as shown in Fig.~\ref{fig:arc}. Specifically, RFMs and the PPM  as the feature encoder transform input images to high-dimensional feature maps. The SPCM and a bilinear upsampling layer decode the high-dimensional feature maps to the high-resolution density maps to achieve crowd counting.

\subsection{Residual fusion module}
\label{sec:dilated}
The major issue is to deal with highly varied sizes of objects, which poses great challenges to regular convolutional networks because they perform feature extraction by sliding fixed-size convolutional kernels on the input feature map. The multi-column architecture is extensively explored which however is computationally expensive and hard to optimize when the network goes deeper. Rather than relying on the multi-column structure, we propose the residual fusion module for multi-scale feature extraction. We introduce dilated convolution to enlarge the reception field which captures contextual information from a larger range comparing to standard convolution \cite{chen2017rethinking,yu2015multi}. Specifically, the RFM is built by integrating convolutional kernels with multiple dilated rates, which establishes nested dilated convolutional layers to extract multi-scale features. To be more precise, we divide the kernels in each nested dilated convolutional layer into $K$ groups, where each group $G_{k}$ uses a dilation rate of $2^{k-1}$ where $k \in {1,2,...,K}$. We would like to highlight that replacing the standard convolution kernels with dilated kernels introduces no additional parameters or computational cost, which makes our network to be computationally affordable.

Moreover, to leverage the effectiveness of the residual learning for effective training without suffering from degradation, we adopt the short-cut connection in our RFMs, which is typically implemented by an identity mapping and an $1 \times 1$ convolutional layer \cite{he2016deep}. We incorporate the short-cut connection to every two nested convolutional layers. In particular, the projection short-cut implemented by an $1 \times 1$ convolutional layer is used for dimension matching when the resolution or the number of channels of feature maps change.

Four nested dilated convolutional layers in conjunction with two short-cut connections constitute a residual fusion module, and four RFMs are stacked for the hierarchical multi-scale feature extraction, with sub-sampling operations by a factor of 2 implemented when features are transmitted between residual fusion modules.

In the RFMs, the feature map has undergone 4 downsampling operations, which means that the hight $h_f$ and width $w_f$ of the final feature map reduce to $\frac{h}{16}$ and $\frac{w}{16}$, respectively, where $h$ and $w$ correspond to the hight and width of the input image, respectively. Considering an input image with large resolution, the corresponding largest valid receptive field still covers a limited spatial area, which might limit the contextual information provided by the receptive fields. To further enhance contextual information aggregation without inducing much additional computation, we introduce the pyramid pooling module (PPM) to efficiently fuse the features at multiple scales \cite{zhao2017pyramid}.

\subsection{Pyramid pooling module}
\label{sec:ppm}


Specifically, given the final feature map of $h_{f} \times w_{f} \times c_{f}$, we apply average pooling operations at multiple scales to aggregate sub-regional contextual information at different scales, where the kernel sizes of the $K$ average pooling layers are $\frac{{{h_f}}}{2^k} \times \frac{{{w_f}}}{2^k}$, and $k = 0,1, \cdots ,K - 1$. After pooling the feature using $K$ kernels and getting the pyramid features of $K$ resolutions, we simply resize them back to $h_{f} \times w_{f}$ using nearest neighbour interpolation, which produces a series of features of the same resolution $h_{f} \times w_{f}$. Together with the original feature, we concatenate them and derive the pyramid feature of size $h_{f} \times w_{f} \times (k+1)c_{f}$. An $1 \times 1$ convolutional layer is followed to aggregate the feature back to $h_{f} \times w_{f} \times c_{f}$. We set $K = 4$ in experiments for a balance between performance and computation.

The PPM further expands the receptive fields to different scales and abstracts the information of sub-regions in different sizes by adopting multi-scale pooling kernels in a few strides. The aggregation of multi-scale contextual information provides more powerful representations to distinguish individuals in different sizes from the background, which can aid our network to make more accurate estimation and get better density maps.

\subsection{Sub-pixel convolutional module}

The convolution operations with pooling layers progressively reduce the feature resolution in exchange for larger valid reception field and invariance. Directly generating a low-resolution density prediction from the final convolutional feature would be not optimal, since the low-resolution prediction results in blurry estimation, which always performs poorly, especially at the points that density varies dramatically. Therefore, it is necessary to generate the density map in the same size as the input images. Deconvolutional layers could be adopted for recovering feature resolution but the considerable computation cost and the difficulty of training make it a sub-optimal choice for crowd counting.

To implement an efficient and easy-to-train feature resolution recovery, we introduce the sub-pixel convolutional module (SPCM) to leverage its great effectiveness in resolution recovery \cite{shi2016real} to crowd counting and density estimation. The SPCM rearranges the elements of the feature map in a size of $\frac{H}{r} \times \frac{W}{r} \times C \cdot r^2$ to a feature map of the shape $H \times W \times C$, which recovers the resolution of the feature maps in a precise way with nearly no computational cost. In practice, the parameter $r$ is set to $4$ in our network, which means that the SPCM increases the spatial resolution by a factor of $4$. Then a bilinear interpolation operation follows to upsample the feature map to the final density map in the same size as the input image.

The SPCM with a following bilinear upsampling layer provides a nonparametric way to recover the resolution of the feature maps and generate the final density map. Moreover, in our work, the rearrangement of the elements in the feature map explicitly guide the network to use the information encoded in the channel dimension to compensate the loss of spatial resolution, so that even though no additional computation cost is introduced, the spatial information can still be well preserved in the channel dimension.

\section{Data Preparation}

We train our network to estimate the density map instead of directly predicting the total amount for crowd counting like most of the recent methods, because density map preserves more information and improves the performance of crowd counting \cite{zhang2016single}.
To further augment the limited training data while maintaining the preciseness of the data, we put forward two principles for data preparation and propose online sampling and multi-scale training following these principles to train our network, which further enhances the robustness and performance of the network.

\subsection{Density map generation}

In existing methods \cite{zhang2016single}, the Gaussian kernel is widely used for density map generation, which specifically puts a Gaussian kernel normalized to $1$ on each of the pedestrian annotations. Based on the generated density maps, data augmentation, e.g. scaling, is usually adopted to generate more training samples. However, this paradigm of data augmentation would induce misleading information. It is obvious that the sizes and peaks of the annotations are only related to the density of pedestrians and independent of the sizes of pedestrians, which means that the information provided by our ground truth should depend only on the crowd density rather than the pedestrian sizes.

To sum up, we come up with two rules for data preparation, which should be followed in density map generation for crowd counting:

\begin{itemize}
	\item Each pedestrian in the image should be annotated only according to the density distribution rather than the size of the object. This is because it is the number of objects that matters in crowd counting rather than the size of the objects.
	
	\item The information provided by the annotation of each object should be consistent during data augmentation, e.g., scaling, that is, the sizes of Gaussians associated with objects should be the same.
\end{itemize}

\begin{figure}[t]
	\begin{center}
		\includegraphics[width=1.0\linewidth]{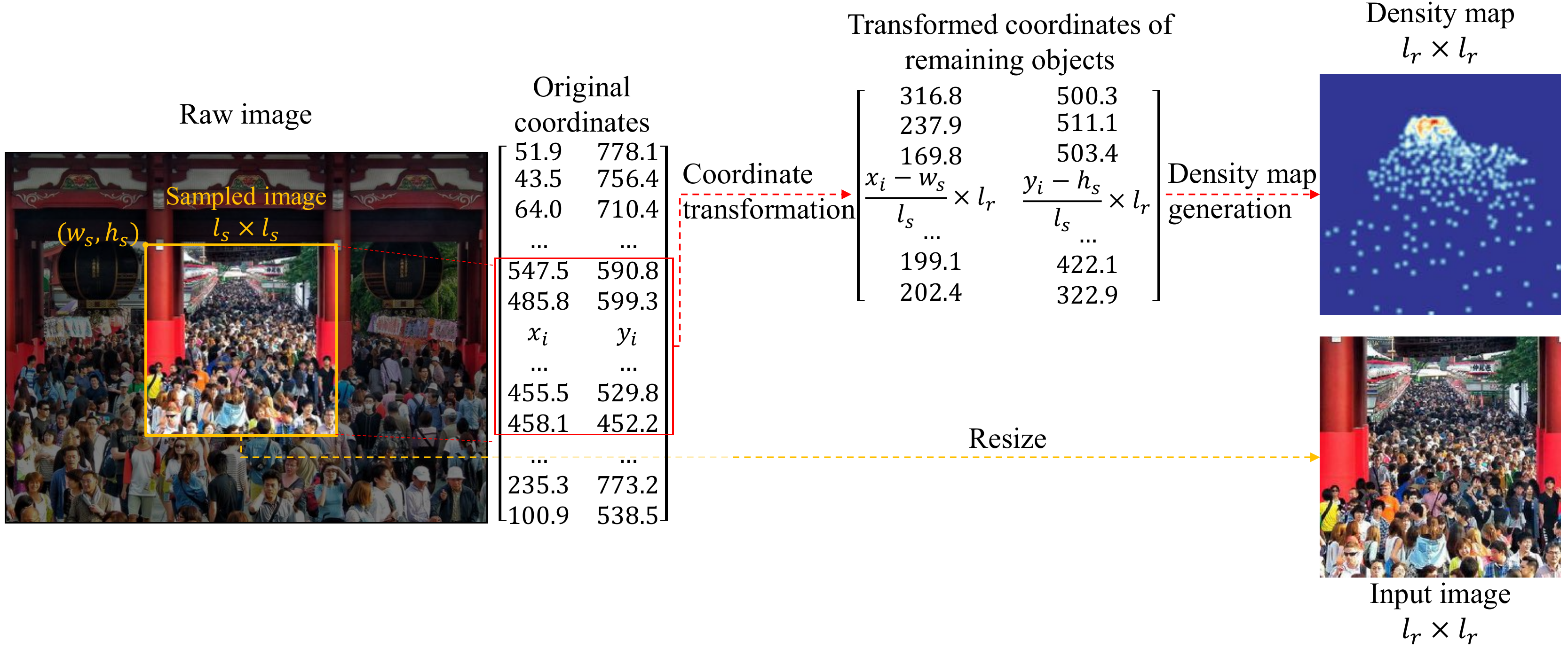}
	\caption{Illustration of online sampling. The online sampling achieves a powerful data augmentation without changing the distribution information in the ground truth density map.}
	\label{fig:data}
	\end{center}
\end{figure}

\subsection{Data augmentation}

As mentioned in previous sections, the robustness to various pedestrian sizes is the key to high-quality crowd counting and density estimation, which needs a large amount of training data and full utilization of the data. However, the training samples of the public datasets are limited due to the heavy cost of data annotation. To further augment the training data and make full use of the data without introducing misleading information in conventional augmentation methods, we propose online sampling and multi-scale training by following the aforementioned rules.

\paragraph{Online sampling.} In existing work, data augmentation, e.g., random cropping and scaling the input image, has been widely used, which is realized by generating the density maps for the original images before training in an off-line way. Accordingly, ground truth density maps are generated based on the augmented data; and then normalization is applied to the resized density map to ensure that the sum of the density map remains unchanged after resizing \cite{zhang2016single}.

However, ground truth generation in this way changes the sizes of the Gaussian kernels in the density map which violates the aforementioned rules.
 Therefore, we propose to use online training sample generation. Specifically, the annotations of the objects are not transformed to density maps before training. During training, we first randomly sample a square area in the size of $l_{s} \times l_{s}$ from the input image with the size of $h \times w$ and scale the sampled image to the size of $l_{r} \times l_{r}$, where $l_{s} \in [\frac{h}{2}, h]$ and we assume that $h \leq w$ for the sake of discussion. Afterwards, we compute the relative coordinates of the objects in the sampled image and put fixed size Gaussian kernels on these coordinates to get the ground truth density map of size $l_{r} \times l_{r}$. The online sampling achieves a powerful data augmentation without changing the distribution information in the density map. An illustration of the online sampling is shown in Fig.~\ref{fig:data}.
\paragraph{Multi-scale training.}
%
%

Since our network is built based on fully convolutional network architecture, it can take inputs of any sizes.  We further increase the randomness by defining multiple parameter $l_{r}$ for augmenting training samples. Specifically, we define a list $L_r$ containing all the candidate $l_r^{i}$, and randomly select one value from the list to be the identical sample size $l_r$ in each iteration. In contrast to existing methods using single-scale training, our multi-scale training introduces much more variations of data, which improves the ability of our network to handle highly dense scenes.

\begin{table}
	\begin{center}
		\small{
			\begin{tabular}{|l|ccccc|}
				\hline
				Dataset & No. of images & No. of train & No. of test & Resolution & Total count \\ \hline
				ShanghaiTech A \cite{zhang2016single} & 482 & 300 & 182 & Varied & 241677 \\ \hline
				ShanghaiTech B \cite{zhang2016single} & 716 & 400 & 316 & 768 * 1024 & 88488 \\ \hline
				UCF\underline{ }CC\underline{ }50 \cite{idrees2013multi} & 50 & 40 & 10 & Varied & 63705 \\ \hline
				WorldExpo'10 \cite{zhang2015cross} & 3980 & 3380 & 600 & 576 * 720 & 199923 \\ \hline
		\end{tabular}}
	\end{center}
	\caption{Summary of the three datasets}
	\label{tab:1}
\end{table}

\section{Experiments}

We conduct extensive experiments on three benchmark datasets, which are widely-used for crowd counting. The statistics of the three datasets, i.e., the ShanghaiTech dataset \cite{zhang2016single}, the UCF\underline{ }CC\underline{ }50 dataset \cite{idrees2013multi} and the WorldExpo'10 dataset \cite{zhang2015cross} are provided in Table 1. We also provide comprehensive comparison with state-of-the-art methods.

\subsection{Evaluation metrics}

Following the traditional protocol of crowd counting works \cite{sindagi2017generating,zhang2016single}, we evaluate all of the recent methods with the Mean Absolute Error (MAE) and the Mean Squared Error (MSE), which are defined as $MAE = \frac{1}{N}\sum\limits_{i = 1}^N {|{C_i} - C_i^{gt}|}$ and $ MSE = \sqrt {\frac{1}{N}\sum\limits_{i = 1}^N {|{C_i} - C_i^{gt}{|^2}} }$, respectively,
where $N$ is the number of images of the test set, and $C_i^{gt}$ and $C_i$ represent the ground truth count and the predicted count of the $i$-th image, which are computed by the sum of the density maps. 

\subsection{Results}
\begin{figure}[t]
	\begin{center}
		\includegraphics[width=1.0\linewidth]{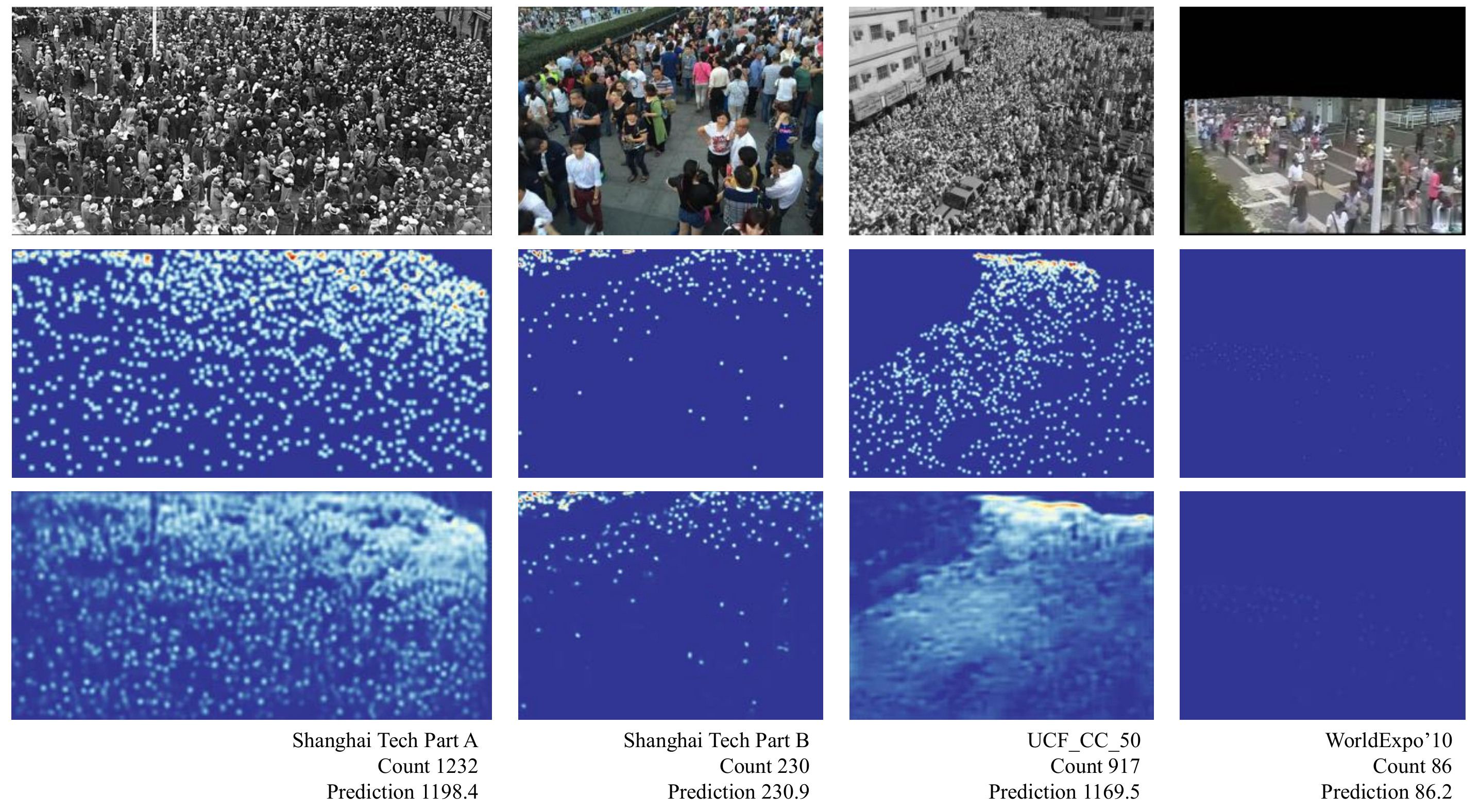}
	\end{center}
	\caption{Illustrative results on three datasets. Original images, ground truth density maps, and the estimations of SCNet are shown from the top row to the bottom row.}
	
	\label{fig:dm}
\end{figure}

We report experimental results on three datasets, and provide comprehensive comparisons with other previous methods. Our SCNet consistently delivers high performances and exceeds recent state-of-the-art methods in most cases, which shows the great effectiveness of our SCNet as a simple but powerful solution for crowd counting. Illustrative results are visualized in Fig~\ref{fig:dm}.

\paragraph{ShanghaiTech.}The ShanghaiTech dataset \cite{zhang2016single} consists of two subsets: Part A and Part B as shown in Table~\ref{tab:1}. We evaluate our network on both subsets. The results and comparison with other methods are reported in Table~\ref{tab:2}. Our SCNet achieves the lowest MAE on Part A among all compared methods; and it produces the lowest errors of both in MAE and MSE on Part B, which dramatically outperforms other methods by large margins.

\begin{table}
\begin{center}
\small
\begin{tabular}{|l|cc|cc|cc|}
\hline
\qquad & \multicolumn{2}{|c|}{ShanghaiTech Part A} & \multicolumn{2}{|c|}{ShanghaiTech Part B} & \multicolumn{2}{|c|}{UCF CC 50}\\ \hline
Method & MAE & MSE & MAE & MSE  & MAE & MSE\\
\hline\hline
Idrees et al. & - & - & - & - & 419.5 & 541.6\\ \hline
Zhang et al. & 181.8 & 277.7 & 32.0 & 49.8 & 467.0 & 498.5 \\
Marsden et al. & 126.5 & 173.5 & 23.8 & 33.1 & 338.6 & 424.5 \\
MCNN & 110.2 & 173.2 & 26.4 & 41.3 & 377.6 & 509.1\\ \hline
Cascaded-MTL & 101.3 & 152.4 & 20.0 & 31.1 & 322.8 & 397.9\\
Switching-CNN & 90.4 & 135.0 & 21.6 & 33.4 & 318.1 & 439.2\\
CP-CNN & 73.6 & \textbf{106.4} & 20.1 & 30.1 & 295.8 & \textbf{320.9}\\ \hline
Ours & \textbf{71.9} & 117.9 & \textbf{9.3} & \textbf{14.4} & \textbf{280.5} & 332.8\\ \hline
\end{tabular}
\end{center}
\caption{Estimation errors on the ShanghaiTech dataset and the UCF\underline{ }CC\underline{ }50 dataset}
\label{tab:2}
{-3mm}
\end{table}

\begin{table}
\begin{center}
\small
\begin{tabular}{|l|c|c|c|c|c|c|}
\hline
Method & Scene1 & Scene2 & Scene3 & Scene4  & Scene5 & Average \\
\hline\hline
Chen et al. & 2.1 & 55.9 & \textbf{9.6} & 11.3 & 3.4 & 16.5\\
Zhang et al. & 9.8 & 14.1 & 14.3 & 22.2 & 3.7 & 12.9\\
MCNN & 3.4 & 20.6 & 12.9 & 13.0 & 8.1 & 11.6\\ \hline
Switching-CNN & 4.4 & 15.7 & 10.0 & 11.0 & 5.9 & 9.4\\
CP-CNN & 2.9 & 14.7 & 10.5 & \textbf{10.4} & 5.8 & 8.86\\
Ours & \textbf{1.8} & \textbf{9.6} & 14.2 & 13.3 & \textbf{3.2} & \textbf{8.4}\\ \hline
\end{tabular}
\end{center}
\caption{The MAE of the WorldExpo'10 dataset}
\label{tab:3}
\end{table}

\paragraph{UCF\underline{ }CC\underline{ }50.} The UCF\underline{ }CC\underline{ }50 dataset \cite{idrees2013multi} contains only 50 images collected from diverse scenes with varying perspective and a wide range of densities, which makes the dataset extremely challenging. To overcome the extremely limited images we perform a 5-fold cross-validation following the standard setting in \cite{idrees2013multi}. The results and comparisons of our method and the other 9 recent methods are shown in Table~\ref{tab:2}. Again, our SCNet achieves the highest performance in terms of MAE with a remarkable improvement over compared methods, and very competitive performance in terms of MSE with the state of the arts.

\paragraph{WorldExpo'10.}Frames of the WorldExpo'10 dataset \cite{zhang2015cross} are from 108 different scenes with annotated ROI for each of the scenes. The training frames and test frames are taken from different 103 scenes and the remaining 5 scenes respectively. We follow the settings in \cite{zhang2015cross} by only considering the ROI regions. The results are shown in Table~\ref{tab:3}. Our SCNet delivers the lowest MAE in 3 of the 5 test scenes, and also achieves the best performance in terms of average MAE.

We have also conducted an ablation study to show the effectiveness of our new paradigm of data preparation. Specifically, we perform experiments on the ShanghaiTech Part B dataset by individually removing online sampling and multi-scale training. The experimental results are reported in Table~\ref{tab:4}, which show that both of the online-sampling and multi-scale training can improve the performance in crowd counting.

\begin{table}
\begin{center}
\small{
\begin{tabular}{|l|cc|}
\hline
Method & MAE & MSE\\
\hline\hline
SCNet & 10.6 & 16.7 \\
SCNet+Online sampling & 9.8 & 15.0 \\
SCNet+Online sampling+Multi-scale training & \textbf{9.3} & \textbf{14.4} \\ \hline
\end{tabular}}
\end{center}
\caption{Estimation errors of different configurations of our methods on ShanghaiTech Part B dataset.}
\label{tab:4}
\end{table}

\section{Conclusion}

In this paper, we have presented the single-column counting network (SCNet) for crowd counting via density estimation. We propose residual fusion module which adopts dilated convolution for feature extraction at multiple scales, and pyramid pooling module for efficient feature aggregation. To balance the accuracy of pixel-wise density estimation and computational cost, an efficient sup-pixel convolutional module is proposed to recover feature resolution and encourage spatial information to be encoded in channel dimension without any parameters. In order to further fully make use of the limited training data, we introduce our training setting empowered by online sampling and multi-scale training, which boosts the robustness of our network and provides a principled setting for crowd counting task. The experiment results on three benchmarks have demonstrated our SCNet as a powerful tool for crowd counting and density estimation.

\section{Acknowledgement}

This paper was supported in part by the National Key Research and Development Program of China under Grant 2016YFB1200100, the National Natural Science Foundation of China under Grant 91538204 and Grant 61425014, the Foundation for Innovative Research Groups of the National Natural Science Foundation of China under Grant 61521091, and National Science Foundation.

\bibliography{egbib}
\end{document}